\documentclass[12pt]{article}

\usepackage{moreverb,url}
\usepackage{gensymb}
\usepackage{lscape}
\usepackage{subfig}
\usepackage{graphicx}
\usepackage{amsmath}
\usepackage{amssymb}
\usepackage[overload]{empheq}
\usepackage{caption}
\usepackage{gensymb}
\usepackage[backend=bibtex,style=authoryear,sorting=none]{biblatex}
\bibliography{references.bib}

\usepackage[colorlinks,bookmarksopen,bookmarksnumbered,citecolor=red,urlcolor=blue]{hyperref}

\title{Decentralized traffic management of autonomous drones}

\begin{document}

\author{
Boldizsár Balázs,$^{1,\ast}$ Tamás Vicsek,$^{1,2}$ Gergő Somorjai$^{2,3}$\\
Tamás Nepusz,$^{3}$ Gábor Vásárhelyi$^{2,3}$\\
\\\normalsize{$^{1}$Department of Biological Physics, Eötvös University of Budapest, Hungary}\\
\normalsize{$^{2}$ MTA-ELTE Statistical and Biological Physics Research Group, Budapest, Hungary}\\
\normalsize{$^{3}$ CollMot Robotics LLC, Budapest, Hungary}\\
\\
\normalsize{$^\ast$To whom correspondence should be addressed;
boldizsar.balazs@ttk.elte.hu}
}


\date{}


\baselineskip24pt


\maketitle


\begin{abstract}
    Coordination of local and global aerial traffic has become a legal and technological bottleneck as the number of unmanned vehicles in the common airspace continues to grow. To meet this challenge, automation and decentralization of control is an unavoidable requirement. In this paper, we present a solution that enables self-organization of cooperating autonomous agents into an effective traffic flow state in which the common aerial coordination task - filled with conflicts - is resolved. Using realistic simulations, we show that our algorithm is safe, efficient, and scalable regarding the number of drones and their speed range, while it can also handle heterogeneous agents and even pairwise priorities between them. The algorithm works in any sparse or dense traffic scenario in two dimensions and can be made increasingly efficient by a layered flight space structure in three dimensions. To support the feasibility of our solution, we experimentally demonstrate coordinated aerial traffic of 100 autonomous drones within a circular area with a radius of 125 meters.
    
\end{abstract}

\section{Introduction}

With the rapid growth rate of drone use today the demand for safe and reliable Unmanned Aircraft Systems Traffic Management (UTM) solutions has also increased critically \parencite{SydAli2019TrafficCity}. In the US the FAA currently registers around 200 000 manned aircraft generating only a couple of thousand flights simultaneously at peak times, while there are already over half million recreational and over 300 000 commercial drones registered with over 240 000 pilots who wish to use the same airspace as flexibly and as frequently as possible \parencite{FederalAviationAdministrationAirNumbers}. (See link, \href{https://www.faa.gov/air_traffic/by_the_numbers/}{here}.) 
It is foreseeable that handling unmanned aircraft as part of the airspace traffic management system will require unprecedented levels of automation and decentralization with proper path planning and detect-and-avoid (DAA) solutions that take the burden off the shoulders of both air traffic controllers and pilots, providing levels of efficiency, safety and reliability that would not even be possible by pure human control.

The density of autonomous or semi-autonomous unmanned aerial vehicles will increase even more in future smart cities where drone services for inspection, safety monitoring or delivery of goods and people will be part of everyday life, and thus drone traffic control will become a minimal requirement of drone usage \parencite{Khan2020EmergingChallenges, Foina2016DronesResearch}.

Providing a full solution to the UTM integration challenge is an extremely complex, long-term process that requires cooperation from many disciplines, stakeholders, authorities and users \parencite{Rumba2020TheTraffic-management}. In this paper we concentrate on a very specific aspect of UTM: we demonstrate the decentralized resolution of complex traffic scenarios generated by high numbers of autonomous vehicles. In other words, we deal with the challenges of large scale open-air drone traffic that arise in an emergent manner when the number and the density of agents and thus the conflicts between them increase by orders of magnitude.

The situation evokes the early days of 'standard traffic', i.e., when cars traveling on the streets first experienced the reduction of the free space between them to a degree, which prevented free flow and resulted in a loss of efficiency in reaching their destination \parencite{Greenshields1935ACapacity}. To avoid traffic bottlenecks, several major structural improvements have been invented and introduced. These included increasing the number of lanes both along the roads and at the cross-sections and other locations relevant to smooth flow. An alternative method was the introduction of one-way streets improving the overall throughput of a network of streets by providing more space for cars moving towards pre-defined destinations. Following some prior works, in the early 1990s several groups of physicists developed theoretical approaches to the quantitative description and prediction of phenomena occurring during traffic situations, both involving vehicles and pedestrians. Most of these works is presented in the review by Helbing \parencite{Helbing2001TrafficSystems}.

One of the essential features of vehicular traffic is that despite the above improvements, it is still confined to a quasi-one-dimensional area: the roads/streets. The main reason for this is that efficient traffic requires a good quality surface, while producing such surfaces is costly. As a result, all sorts of cars are moving along a network of roads with two or more lanes, where the lanes are assigned to a unique direction (one out of two), while cross-sections are managed by streetlamps or signs/rules determining the behavior of a vehicle arriving at a cross-section or roundabout.

The traffic situation involving aerial units is facing similar challenges in many aspects as the density of aircraft is increasing. The standard air traffic control response to this is restrictive, as it builds on the well-known concepts of road traffic: the airspace is increasingly structured, quasi-one dimensional virtual lanes are created for each vehicle in it, the general right-hand rule is used that results in virtual roundabouts etc. On the other hand, the essence of airspace seems different from road infrastructure, primarily because there is no extra cost for using extra space. In other words, changing the actual path (e.g., to avoid another aircraft) does not require expensive additional road infrastructure. This allows for much more freedom in the conflict resolution between flying agents compared to what we are used to on the ground, but this freedom can be exploited only with a very different attitude that involves decentralization, decrease of "road" structure and increase of local problem solving.

There have been many recent attempts to provide solutions for the coordination of multiple aerial vehicles. Centralized collision-free multi-robot path planning is a major direction where the NP-hard mathematical problem of generating non-overlapping 4D trajectories is solved by a central computer before takeoff or during flight. Used methods include numerical approaches, such as mixed integer linear programming \parencite{Grtli2012PathMILP}, sequential convex programming \parencite{Augugliaro2012GenerationApproach}, graph search \parencite{Jose2016TaskMethods} or GPU-optimized gradient descent \parencite{Hamer2019FastAcceleration}, bio-inspired methods such as ant colony optimization \parencite{Wu2021Swarm-BasedEnvironments} or particle swarm optimization \parencite{Duan2013HybridReconfiguration}, and many others \parencite{Zhou2020UAVTrends, Maity2021FlyingTrends, Madridano2021TrajectoryApplications, Mellinger2011MinimumQuadrotors}.

To the best of our knowledge, the largest UAV swarm up to now with real-time central control for collision-free path planning is presented in \parencite{Luis2020OnlinePlanning} with 20 miniature drones; it uses an algorithm called Distributed Model Predictive Control, works indoors and relies on a high-end motion capture system to acquire the accurate position of the agents. The downside of central solutions -- let them be a priori or real time -- is that they have single points of failure, require very stable communication infrastructure and they are less scalable, generally less adaptive and more bureaucratic than anything calculated on the fly in a distributed manner.

Decentralized multi-UAV path-planning methods are also getting increasingly popular. \parencite{Zhou2021EGO-Swarm:Environments} and \parencite{Dmytruk2021SafeEnvironments} both show an autonomous and decentralized solution for multi-UAV navigation in cluttered environments using up to three real drones. \parencite{Quan2021SkyTraffic} shows an interesting design idea of geometric aerial highways that helps flying agents get organized into road-like lanes in the sky, however, this solution also involves severe restrictions on the free motion of the agents. \parencite{Zaini2020DistributedCommunication} demonstrates the advantages of a distributed, decentralized traffic coordination using only low-bandwidth communication, however presents only simulation results and no actual experiments on real hardware. \parencite{Zehavi2021HybridManagement} suggests a hybrid method for centralized global and decentralized local UTM. 

Going beyond algorithm design or simulations and showing actual real-world experiments with drone swarms is much harder than one thinks. There is only a very limited number of groups, that we know of, who could actually demonstrate the real-time outdoor coordination of drones with more than a few flying agents. Contrarily, the largest drone displays by now consist of more than 5000 drones \parencite{HighGreat2021ChineseShow}, but these entertaining flights are pre-programmed and contain no real-time conflict resolution. The largest autonomous drone swarms presented so far perform flocking or formation flights, i.e., missions where the task is to synchronize motion and move together
\parencite{Schilling2021Vision-BasedEnvironments}. The largest of these we know of is our latest solution with 52 drones \parencite{Balazs2020AdaptiveFlocking}. 

Autonomous traffic differs from flocking in the sense that the task is completely different: instead of moving together, agents need to move on their individual paths, but in a coordinated manner. These two - moving together in a flock and moving independently in traffic - provide two of the main building blocks of coordination upon which most of the multi-UAV missions can be based. Regarding drone traffic, an interesting paper is given by \parencite{Leven2011DealingSystems} with 5 fixed-winged aircraft that avoid conflicts by "altitude differentiation". Large-scale demonstration of traffic coordination of UAVs include our own previous publication with 30 drones \parencite{Balazs2018CoordinatedDrones} which we exceed substantially in this paper, and an US Air Force marketing video with 100 miniature fixed-wing aircraft which move together from point A to point B and loiter there \parencite{USAirForcePerdix}. Unfortunately, this video provides no actual details on the coordination or control aspects and thus cannot be treated as a scientific result.

A major goal in both the observational and theoretical approaches to traffic is the determination of the so-called fundamental diagram \parencite{helbing2009derivation}, which describes perhaps the most important feature of a traffic situation: 
the dependence of the average flow of vehicles as a function of their average density. 
The corresponding plot starts out quite trivially, since – assuming a constant average speed in the absence of disturbances – the global flow is linearly growing with the density of vehicles for its small values. 
The situation gradually changes because keeping a safe distance from others increasingly slows down the flow because of the non-zero reaction time of the drivers as well as their stopping distance. The fluctuation of these parameters results in instabilities and eventually results in all sorts of traffic phenomena including jamming, stop-and-go waves, etc. \parencite{Helbing2001TrafficSystems}. Such phenomena occur even along a given long road and the effects due to networked roads adds further difficulties to making the flow of vehicles efficient \parencite{Mollier2018AMethods, Zeng2020MultipleTraffic}.             
Interestingly enough, similar kinds of fundamental diagrams can also be constructed for pedestrian traffic \parencite{Vanumu2017FundamentalReview}. Although pedestrians  – quite like the way drones fly in a 2D layer – can, in principle, easily avoid collisions and get around each other by stepping aside while walking (they do not have to follow strictly defined, one way lanes), at high densities jamming and even disastrous accumulation of pressure within the crowd can occur, leading to even death of panicking people \parencite{Helbing2000SimulatingPanic}. Although in this case an extra dimension is provided for avoiding each other, the joint effect of a large density and a desired destination results in difficulties in maintaining a seamless flow. In part because of this, we expect that the flight of drones in layers assigned for them for traffic (which is a common practice for aerial vehicles) does not automatically guarantee that they can easily move towards their destinations.

The main contributions of this paper regarding autonomous drone traffic are summarized below. We present a completely decentralized solution for unmanned traffic management with a combined method of path planning and detect-and-avoid that provides unprecedented efficiency even in extremely dense and complex traffic scenarios. We designed our algorithms to work in two dimensions, as a realistic scenario for horizontal traffic, however, we also present how layered traffic in three dimensions can maximize the traffic flow. We show that our solution is scalable both in the number of agents and in the velocity space of agents, and can also effectively handle priorities and hierarchies among the agents. The solution is presented in simulation with up to 5000 agents in very crowded traffic scenarios and also with 100 autonomous drones out on the field. 

Our solution builds on bio-inspiration in many aspects. First of all, we provide an agent-based velocity controller for traffic that incorporates the most important interaction terms of general flocking models describing the collective motion of animal herds (and pedestrians): repulsion and alignment \parencite{Vicsek1995}. This allows us to use a completely decentralized mindset which is ruled by simple self-propelling terms and pairwise interactions, which gives rise to truly scalable systems (nevertheless, we emphasise again that a flocking task is fundamentally different from a traffic task, and the usage of flocking solutions for traffic is limited to its concept and to some modified interaction terms). Second, we use evolutionary optimization \parencite{Eiben2015IntroductionAlgorithms} to tune the numerous control parameters of the agent-based model in a realistic simulation framework similarly as in our previous work \parencite{Vasarhelyi2018}. Third, we also deal with the structural requirements of large-scale natural systems \parencite{Anna2018}: we incorporate heterogeneity and allow hierarchical priority structures in our algorithms.

Throughout this paper we will be considering general traffic scenarios among agents whose flight characteristics are omnidirectional and their speed range includes hovering (such as multirotor drones). These agents will be tasked to reach their independently assigned destinations while avoiding each other in open space.

\section*{Traffic algorithm}

Our algorithm for decentralized aerial traffic is the sum of: i) a \emph{conflict avoidance strategy} with low computational demand fitted to current on-board drone hardware capacities, and ii) an instinct-like \emph{sense-and-avoid behaviour} based on terms migrated from autonomous artificial flocking with reassuring success \parencite{Vasarhelyi2018}.

The on-board \emph{conflict avoidance strategy} calculates a momentarily optimal velocity vector which does not intersect with the projected trajectories of the local neighbours within a reasonable time frame, yet drives the agent efficiently towards its destination. We call this vector the "self-drive" velocity term ($\mathbf{v}_i^\mathrm{self-drive}$ in Eq. \ref{eq_desired-velocity}).
Due to algorithmic simplicity, this momentary desired velocity term can be recalculated at high frequency, so the algorithm is less prone to ill-projected trajectories than more complicated, hence slower, methods.
Regarding its concept, our approach is akin to the family of 'Velocity Obstacle' (VO) based traffic solutions \parencite{VanBerg2008ReciprocalNavigation, VanDenBerg2011ReciprocalAvoidance}. Our main advance compared to previous work presenting centrally computed simulations only is that the VO concept is now suited for the physical reality of outdoor drones and the difficulties of a fundamentally decentralized system.

It is important to note that despite being completely decentralized, our algorithm should optimize each agent's own decision in an altruistic way, according to the group interest within each agents' situation awareness. 
In other words, being decentralized and independent in our case does not lead to individualistic or opportunistic behavior but, on the contrary, to both individual and group benefits, as long as the common rules are followed by everyone.

The \emph{sense-and-avoid method} consists of pairwise repulsive and velocity-alignment interactions ($\mathbf{v}_i^\mathrm{rep}+\mathbf{v}_i^\mathrm{friction}$ in Eq. \ref{eq_desired-velocity}) based on our previously published traffic algorithm \parencite{Balazs2018CoordinatedDrones}, but is now significantly enhanced in terms of effectiveness (almost double-fold increase of traffic throughput in certain situations) and is also generalized to handle hierarchical priorities between agents.

The properly balanced cooperation between the conflict avoidance strategy (planning) and the sense-and-avoid mechanism (akin to reflexes of animals) is crucial in our work. Agents do not switch discretely between these two governing rules as operational states, they rather combine them continuously and adaptively in a symbiotic way. The way agents choose their current velocity to \textit{avoid} conflicts tacitly presumes that sense-and-avoid interactions should be able to \textit{resolve} conflicts if they arise somehow. 
In other words, self-drive is neither too cautious, nor does it have to be absolutely perfect, as it can count on the corrections of the sense-and-avoid mechanism. In the meantime, planning to avoid conflicts with self-drive significantly reduces the load on sense-and-avoid interactions, as dangerously close situations become very rare.

\subsection*{Individual behaviour in decentralized traffic}

Each drone broadcasts its own position, velocity and desired destination with a reasonably high frequency (multiple times per second) to its neighbors, and receives the same data from all of them (see the \textit{Proof of concept with a drone swarm} section for details about the communication network). Our agent-based control algorithm -- executed on all drones in a distributed way -- uses these inputs to determine a momentary desired velocity: $\tilde{\mathbf{v}}_i (t)$, where $i$ denotes the agent. The algorithm is built up of three terms: repulsion, friction and self-drive.

\begin{equation}
    \tilde{\mathbf{v}}_i=\mathbf{v}_i^\mathrm{rep}+\mathbf{v}_i^\mathrm{friction}+\mathbf{v}_i^\mathrm{self-drive}
    \label{eq_desired-velocity}
\end{equation}

The two pairwise interactions, repulsion and friction yield the instinct-like \emph{sense-and-avoid behaviour}. The common form of both repulsion and friction -- that facilitates many robots to flock together in unison \parencite{Vasarhelyi2018} -- was suited before for the quite differing task of traffic where agents have arbitrary individual directions of velocity \parencite{Balazs2018CoordinatedDrones}.

\subsection*{Repulsion} 

Repulsion pushes agents away if they get too close.

\begin{equation}
    \mathbf{v}_i^\mathrm{rep}=\sum_{d_{ij}<R_0} p^\mathrm{rep}\left(R_0-d_{ij}\right)\cdot \hat{\mathbf{u}}_{ij}^\mathrm{rep},
    \label{eq_repsum}
\end{equation}

where $p^\mathrm{rep}$ is a linear gain, $R_0$ is the radius of the repulsive zone, $d_{ij} = |\mathbf{r}_{ij}| = |\mathbf{r}_j - \mathbf{r}_i|$, ($\mathbf{r}_k$ being the position of agent $k$) and $\hat{\mathbf{u}}_{ij}^\mathrm{rep}$ is the direction of this interaction.

Repulsion often pushes agents away from their target, significantly reducing traffic-efficiency and even potentially creating global oscillations. 
Both effects can be mitigated by rotating the direction of repulsion, $\hat{\mathbf{u}}_{ij}^\mathrm{rep}$, a bit away from the obvious (isotropic) direction of $-\mathbf{r}_{ij}/|\mathbf{r}_{ij}|$. 
The direction of this rotation depends on whether the velocities of two agents are roughly the same or not. In the case of agents moving in the opposite direction, it is better to evade each other (i.e. increase the distance mainly on the axis perpendicular to the axis of motion), while in the case of almost parallel velocity vectors, it seems favourable to form lanes \parencite{Helbing2001Self-organizingMovement} (i.e. increase the distance mainly along the axis of motion).

For determining the direction of repulsion, we call $\phi$ the angle between $\mathbf{r}_{ij}$ and the target direction of agent $i$ ($\hat{\mathbf{r}}_{ti}$), and we call $\rho$ the angle between $\hat{\mathbf{u}}_{ij}^\mathrm{rep}$ and the opposite of target direction ($-\hat{\mathbf{r}}_{ti}$). 
In case of isotropic repulsion, commonly used in flocking models: $\rho = \phi$.
Yet in case of anisotropic repulsion:

\begin{equation}
    \rho=
    \begin{cases}
    (1-A)\cdot\phi & \phi\leq\frac{\pi}{2} \text{ and }\mathbf{v}_j \uparrow \uparrow \hat{\mathbf{r}}_{ti}\\
    \pi+(1-A)\cdot\left(\phi-\pi\right) & \phi>\frac{\pi}{2} \text{ and }\mathbf{v}_j \uparrow \uparrow \hat{\mathbf{r}}_{ti}\\
    (1-\frac{A}{2})(\phi-\pi)+\pi & \mathbf{v}_j \uparrow \downarrow \hat{\mathbf{r}}_{ti},
    \end{cases}
    \label{eq_repaniso}
\end{equation}

where $\mathbf{v}_j \uparrow \uparrow \hat{\mathbf{r}}_{ti}$ means less than $\pi/3$ angular difference between the two vectors, $\mathbf{v}_j \uparrow \downarrow \hat{\mathbf{r}}_{ti}$ otherwise.

Numerous graphics and illustrations of the behaviour of 'anisotropic' repulsion can be found in \parencite{Balazs2018CoordinatedDrones}. Throughout the simulations of this paper, we set the level of anistropy, $A \in [0,1]$, to 0.42 as a result of previous evolutionary optimization processes and used the rounded value of 0.5 in real experiments.

\subsection*{Friction}

This interaction -- mimicking the effect of friction between physically touching objects for agents distant from each other -- reduces the velocity difference between close-by agents. Doing so, it effectively increases the safety of traffic by relaxing oscillations (arising in crowded situations when agents get into the repulsive zone of each other) and dampening chaotic dynamics (induced by delays and slow reaction times) \parencite{Orosz2009ExcitingHighways}.

\begin{equation}
    \mathbf{v}_i^\mathrm{friction}=
    \sum_{j\in \mathbb{A}_i}
    \theta\left(v_{ji} - v^{\mathrm{frictmax}} (d_{ij}) \right) \hat{\mathbf{v}}_{ji},
    \label{eq_frictsum}
\end{equation}

    where $\theta$ is the Heaviside step function, $\mathbb{A}_i$ is a set of neighboring agents in relation to whom friction is needed (see the criteria later in the section), unit vector $\hat{\mathbf{v}}_{ji}$ is the direction of agent $j$'s velocity relative to agent $i$'s, and $v_{ij}$ is its magnitude, while $v^{\mathrm{frictmax}}$ is given by

\begin{equation}
    v^{\mathrm{frictmax}}(d_{ij})=
    \mathrm{max} \left(v^\mathrm{friction}, D\left(d_{ij}, R^\mathrm{friction}, p^\mathrm{friction}, a^\mathrm{friction}\right)\right),
    \label{eq_frictmax}
\end{equation}

where $D$ is the smooth braking curve to decay velocity as a function of distance $d$, while explicitly taking into account the motion constraint of finite acceleration capabilities:

\begin{equation}
    D\left(d, R, p, a\right) = 
    \begin{cases}
        0 & \quad d - R < 0 \\
        p(d-R) & 0\leq d-R \leq \frac{a}{p^2} \\
        \sqrt{2a(d-R)-\frac{a^2}{p^2}} & \frac{a}{p^2}<d-R,
    \end{cases}
    \label{eq_linsqrt}
\end{equation}

where $a$ is the acceleration limit, $p$ is a linear gain (to avoid close to infinite derivative of the square root function) in the v-d plane and $R$ is a simple offset value that shifts the $D(.)$ transfer function along its $d$ axis.

Note that a constant deceleration yields a square-root function on the velocity-position plane. With the shape of Eq. \ref{eq_linsqrt}, we force the agents to reduce their velocity-differences with a constant deceleration (of $a^\mathrm{friction}$) and stop (relative to each other) at $R^\mathrm{friction}$ distance from each other. Of course, relative stopping would unnecessarily keep the two agents in close proximity, so we allow a small $v^\mathrm{friction}$ velocity difference at any distance.

Friction of this form has proven to be able to synchronize a large flock of aerial robots outdoors \parencite{Vasarhelyi2018}. But in flocking, aligning velocities is the overall task, while the task of traffic requires differing velocities. 
Thus, in traffic, agent $i$ should not apply friction to each neighbour's current velocity (i.e. the set $\mathbb{A}_i$ in Eq. \ref{eq_frictsum} does not include every agent within the communication range), but select them based on three criteria. 
The first two are related to whether the neighbour poses any danger of collision. It is considered to do so if it comes towards agent $i$ and is lying between agent $i$ and its target. If both hold, friction should be applied. If none, it shouldn't. If one of them, then agent $i$ considers a third criteria related to efficiency. If the neighbour's velocity would drive the agent away from its goal, friction is not applied, otherwise it is. 
Intuitively: selective friction means that agent $i$ intents to align its velocity of agent $j$ if agent $j$ poses a big threat of collision or poses a moderate threat of collision and aligning to it does not mean giving up too much of efficiency.

The above three criteria are defined with fixed angular thresholds:  
'coming towards agent $i$' is checked with a fixed angular threshold of $\pm \pi/4$ between $\mathbf{r}_{ij}$ and $\mathbf{v}_j$; 
'lying between agent $i$ and its target' is checked with a fixed angular threshold of $\pm 2\pi/3$ between $\mathbf{r}_{ij}$ and agent $i$'s target direction; 
'neighbor's velocity would drive agent away from its goal' is checked with a fixed angular threshold of $\pm \pi/2$ between $\mathbf{v}_j$ and agent $i$'s target direction. 
Illustrations of these criteria can be found in \parencite{Balazs2018CoordinatedDrones}.

\subsection*{Self-drive}

\begin{figure}[!htb]
    \centering
    \includegraphics[width=0.9\textwidth]{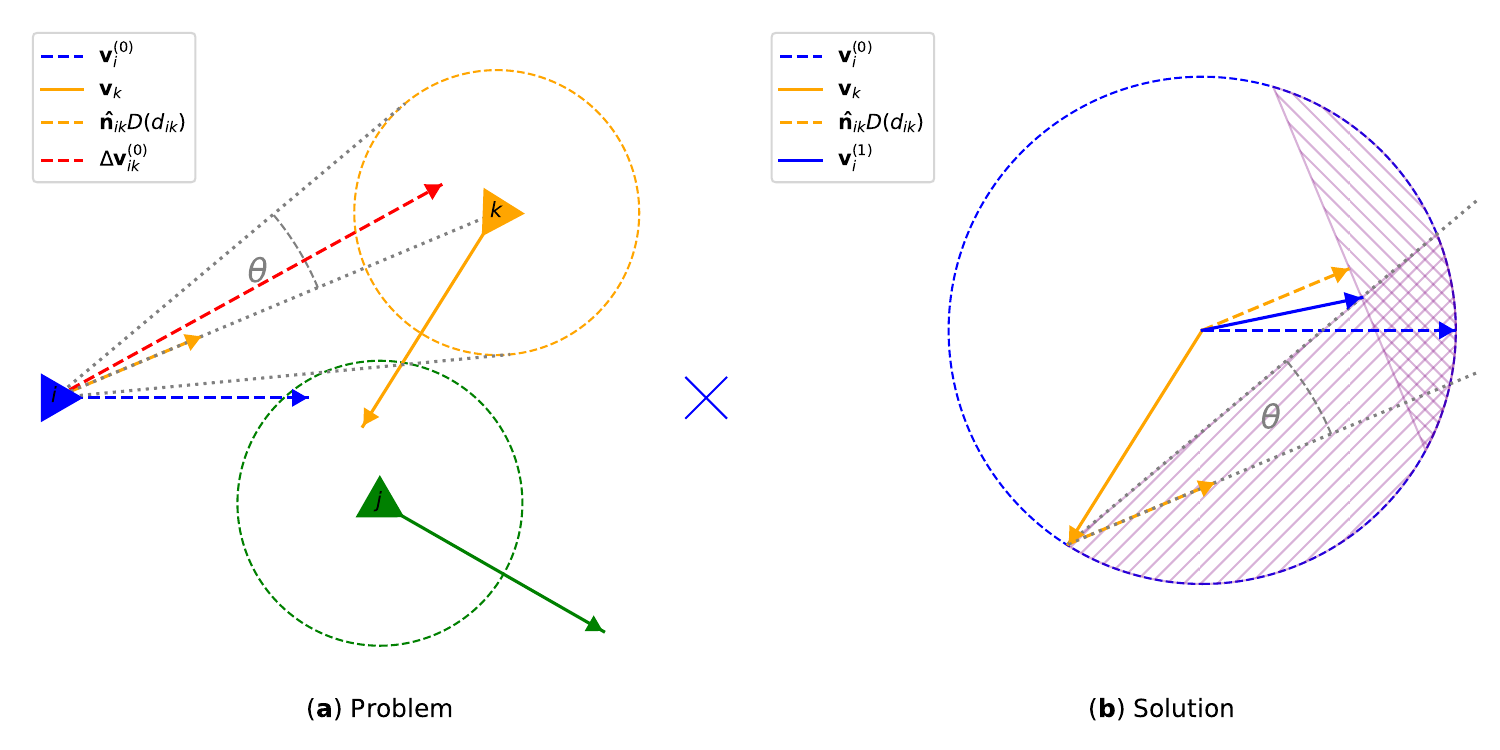}
    \caption[The self-driving term]{The self-driving term. In (a): position and velocity of the agents.
    The linearly extrapolated route of the blue $i$th agent towards its target (blue X) would not intersect with the \emph{current} $R^\mathrm{avoid}$ avoidance area (dashed circle) of the orange $k$th neighbour. Yet it is a threat, because 
    the relative velocity (red dashed vector) would drive the two too close (D1),
    and the projection of the relative velocity to the line between the two is bigger than what $D(d_{ik}, R^\mathrm{avoid}, p^{\mathrm{target}}, a^{\mathrm{target}})$ allows (orange dashed vector, D2). 
    Since the blue agent would partly move towards the orange neighbour (D3)
    and is not going to reach its target before the problem escalates (D4), 
    it is obligated to change its velocity.
    In contrary: despite the projected intersection with its current $R^\mathrm{avoid}$ avoidance area the green neighbour holds no threat thanks to its velocity.
    (b) shows the velocity plane and illustrates how blue agent chooses a new target velocity (solid blue) instead of the rejected one (dashed blue) owing to the orange neighbour's position and velocity.
    The new velocity needs to be outside of the two hatched purple areas (C1-2 and C4) and inside of the blue circle (C3). 
    The choice from the allowed (white) region is the one that has the biggest component towards the rejected velocity (C5). 
    The two purple areas represent two different types of dangers to avoid. 
    Any velocity of the purple "//" hatched area would \emph{either} lead to relative velocity that intersects (C1) with the $R^{\mathrm{avoid}}$ neighbourhood of the orange agent 
    \emph{or} make the blue agent pass on the other side of the orange neighbour as the original velocity would have made it to (C2).
    Velocities in the purple "\textbackslash \textbackslash" hatched area have bigger components towards the orange agent than it would be safe at the current distance (C4).}
    \label{fig_traff_prud-relvel}
\end{figure}

Self-driving is the key of our current algorithm. It should guide the agents towards their individual goal without propelling them too close to each other. If the self-driving terms of close-by agents do mutually well, the instinct-like sense-and-avoid interactions are barely in action.

With the self-driving term, we do not aim to find a mathematically optimal path. It would take too much computation time with the hardware available for drones on-board (solving traffic is an NP-hard problem), and in outdoor flights there are many reasons (wind, sensory noise, movement inaccuracy, temporal loss of inter-agent communication, unexpected modification of the desired missions) why a pre-planned route can fail.
Instead, we construct a set of simple rules to determine a desired momentary self-drive velocity, which takes into account the movement of nearby agents, but doesn't plan too much ahead as those agents themselves can alter their routes in time. Due to its simplicity, the self-drive velocity can be calculated with high frequency in the following fashion.

Agents determine their momentary self-drive velocity ($\mathbf{v}_i^\mathrm{self-drive}$ of Eq. \ref{eq_desired-velocity}) with an iterative method, the method following now. 
For these iterations, $\mathbf{v}_i^{(s)}$ denotes the \textit{candidate} self-drive velocity of the $i$th agent in iteration-step $s$, while $t_{\mathrm{plan}}^{(s)}$ denotes the time until we plan ahead in iteration-step $s$. Initially (in the zeroth step, i.e. with $s$=0)  $\mathbf{v}_i^{(0)}$ points towards the target, and it has a magnitude of $v^{\mathrm{SPP}}$, while we initialize $t_{\mathrm{plan}}^{(0)}$ with the time the agent would get to its target.

In each iteration-step we call a neighbor \textit{threatening} if it satisfies \emph{all} the danger criteria below at once:

\begin{itemize}
\item [D1.]
  the two of them are on a collision course with respect to an avoidance area with radius $R^\mathrm{avoid}$. This is the case, if the
  angle of the relative velocity $\Delta\mathbf{v}_{ij}^{\left(s\right)}\equiv\mathbf{v}_i^{\left(s\right)}-\mathbf{v}_j$  and the relative
  position $\mathbf{r}_{ji}\equiv\mathbf{r}_j-\mathbf{r}_i\equiv d_{ji}\hat{\mathbf{n}}_{ji}$  is smaller than $\text{asin}(\min(R^\mathrm{avoid}/d_{ji},1))$ (noted as $\theta$ on Fig. \ref{fig_traff_prud-relvel}),
\item [D2.]
  the relative velocity is bigger than what the agent could handle on its own
  safely ~~  $\Delta\mathbf{v}_{ij}^{\left(s\right)}\hat{\mathbf{n}}_{ji}>D\left(d_{ji\ },R^\mathrm{avoid},p^\mathrm{avoid},a^\mathrm{avoid}\right)\ $ ,
\item [D3.]
  this undesired situation can at least partly be attributed to
  the~\emph{i}th agent, i.e. it does move towards the other agent to some extent ~ ~ $\mathbf{v}_i^{\left(s\right)}\hat{\mathbf{n}}_{ji}>0$ ,
\item [D4.]
  the time they get too close to each other~ $\left(d_{ji}-r^\mathrm{avoid}\right) / \left|\Delta\mathbf{v}_{ij}\right|$  is
  sooner than~  $t_{\mathrm{plan}}^{\left(s\right)}$ .
\end{itemize}

The final acceptable velocity output of the iterative method ($\mathbf{v}_i^{\mathrm{self-drive}} = \mathbf{v}_i^{(s_\mathrm{last})}$) gets settled when there are no more threatening neighbors.

Otherwise, a new candidate velocity  $\mathbf{v}_i^{\left(s+1\right)}$ is to be calculated. 
If there are more than one threatening neighbours, we deal with the most urgent threat: the neighbour which would get too close first. Let this neighbour be the $k$th agent (for example: on Fig. \ref{fig_traff_prud-relvel}a the orange neighbour is the only threatening one).

The following criteria define the candidate target velocity of the next iteration $\mathbf{v}_i^{\left(s+1\right)}$:

\begin{itemize}
\item [C1.]
  $\Delta\mathbf{v}_{ik}^{\left(s+1\right)}$ needs to be outside of
  the area between the two tangent lines drawn to
  the $R^\mathrm{avoid}$  radius circle around $\mathbf{r}_k$, i.e. the new relative velocity must not yield a collision course,
\item [C2.]
  $\Delta\mathbf{v}_{ik}^{\left(s+1\right)}$ needs to be on the same side of the line defined by $\hat{\mathbf{n}}_{ki}$  as  $\Delta\mathbf{v}_{ik}^{\left(s\right)}$, for not shifting the side on which agent $i$ passes agent $k$ 
\item [C3.]
  $\left|\mathbf{v}_i^{\left(s+1\right)}\right|\leq\left|\mathbf{v}_i^{\left(s\right)}\right|$ , to avoid speeding up in dangerous situations,
\item [C4.]
  $\mathbf{v}_{i}^{\left(s+1\right)}\hat{\mathbf{n}}_{ki}\le D\left(d_{ki\ },R^\mathrm{avoid},p^\mathrm{avoid},a^\mathrm{avoid}\right)$ , to be able to stop before a possible collision,
\item [C5.]
  $\mathbf{v}_i^{\left(s+1\right)} \cdot \mathbf{v}_i^{\left(s\right)}$ must be maximal for increasing overall traffic efficiency.
\end{itemize}


In the situation presented in Fig. \ref{fig_traff_prud-relvel}a the geometrical application of these criteria can be seen on Fig. \ref{fig_traff_prud-relvel}b. 

The velocity $\mathbf{v}_i^{\left(s+1\right)}$ chosen according to C1-C5 is now fed back to the beginning of the iteration step, e.g. checked again by D1-D4. If it does not meet all the danger criteria, it is the choice for $\mathbf{v}_i^{\mathrm{self-drive}}$. Otherwise, the iteration continues with the next $s+1$ step, resulting in $\mathbf{v}_i^{\left(s+2\right)}$ according to C1-C5.

Note that turning away from a potential colliding neighbour is a must in our algorithm, while slowing down depends on the details of the conflict. Hence, after a few iterations it is possible that $\mathbf{v}_i^{\left(s\right)}$ is slower than $v^{\mathrm{SPP}}$, and it is not even moving the agent towards its target at all ($(\mathbf{x}_i^{\mathrm{target}}-\mathbf{r}_i)\mathbf{v}_i^{\left(s\right)} < 0$).
In this case the agent changes its velocity moving exactly towards its target with the reduced velocity magnitude of $\left|\mathbf{v}_i^{\left(s\right)}\right|$ if this choice does not cause any trouble in the sense of the danger criteria D1-D4 introduced above.

Self-driving ends with slowing down nearby the current target of the agent. It is done according to the function introduced in Eq. \ref{eq_linsqrt}, with parameters: $D(d_{\mathrm{target}}, 0, p^\mathrm{avoid}, a^\mathrm{avoid})$, which yields a smooth deceleration and stopping at the target point.

The new concept here, in the self-driving term compared to \parencite{Balazs2018CoordinatedDrones} is that neighboring agents are treated as moving objects. In our previous work \parencite{Balazs2018CoordinatedDrones}, they were treated as static obstacles, and their movement was handled by the fact that agents calculate their desired velocity with high frequency (around 5 Hz). 
Here, the movement of others is involved explicitly in calculating the desired velocity. Hence less caution is needed (for example, $R^\mathrm{avoid}$ can be lower) while smoother avoiding trajectories are attained. This results in significant improvement in performance as seen in the upcoming 'Traffic simulations' subsection. 
Accounting for the movement of neighbors while acquiring the self-drive velocity means, of course, a more complicated set of rules, but the above iterative method is streamlined enough to run live on the on-board computers of our (or almost any commercially available) drones as seen in our experimental demonstration later in the paper.

\subsection*{Queueing}

Agents may share common target points, which requires patience from them -- instead of rushing towards it and jostling around \parencite{Gershenson2015WhenFaster}. Therefore, we introduce a self-organizing queueing behavior to our algorithm. 

For that, every agent communicates its target point to all the others. Agent $A$ queues up behind agent $B$ if $B$'s target is closer to the target of $A$ than $R^\mathrm{avoid}$, and if agent $B$ is closer to its target than agent $A$ is to its own. 

Queueing up in our 2D traffic case means stopping at a distance from the target of the agent, with that distance being a bit further than the other agent ($B$) is to its target.
For further description of how this works, and results on how it leads to an emergent self-organizing 2D-queue, refer \parencite{Balazs2018CoordinatedDrones}.

\section*{Traffic simulations}

\begin{figure}[!htb]
    \captionsetup[subfigure]{justification=centering}
    \centering
    \subfloat[No-interaction null model]{\includegraphics[scale=0.48]{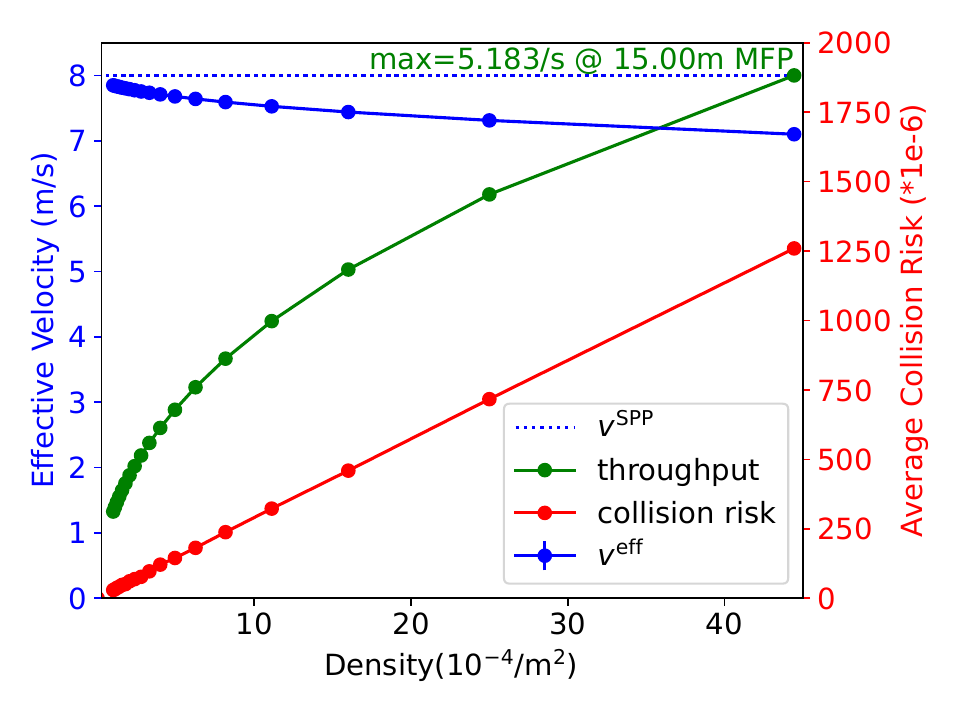}}%
    \subfloat[Optimized traffic algorithm]{\includegraphics[scale=0.48]{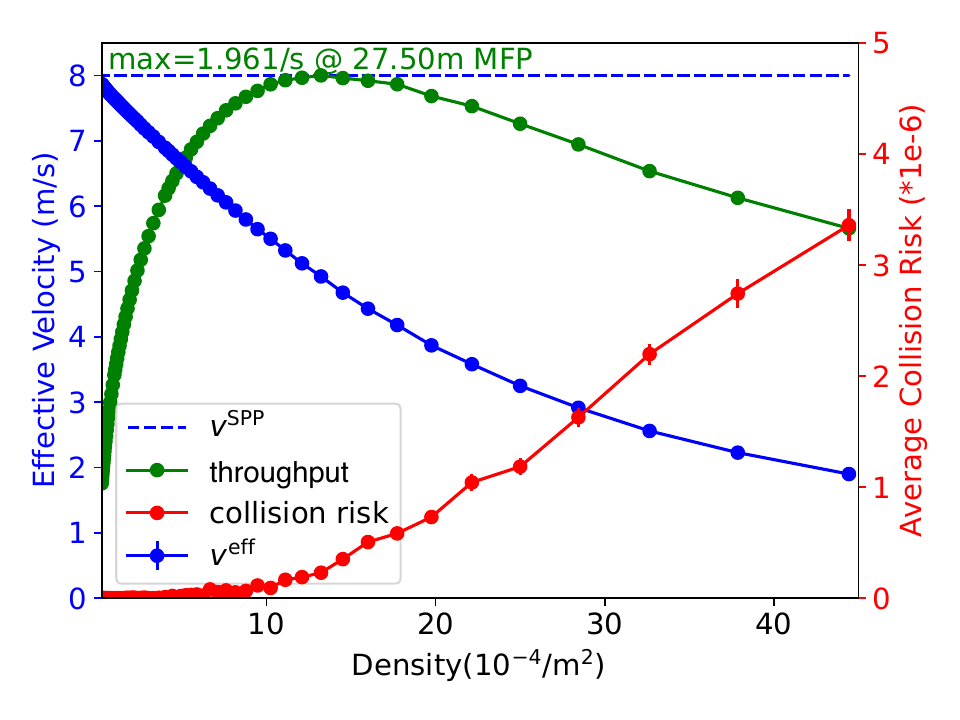}}%
    \caption{Traffic flow statistics of 100 agents as a function of density. Panel (a) is a null model where no agent-agent interaction takes place, while panel (b) shows the optimized traffic flow. One hundred 10-minute simulations were averaged at each measurement point. 
    Standard deviations are shown with error bars, but they are too small for the throughput and the effective velocity to be visible. The throughput with near-to-zero collision risk has a maximum at 27.5 m mean free path ($13.2*10^{-4}$ agents/m$^{2}$), which is a particularly dense traffic situation.}
    \label{fig_traffic-100A-flux}
\end{figure}

The algorithm was studied first in an agent-based simulation framework that has been developed by our group and has been used in most of our previous papers related to swarms of drones. The framework implements a realistic but fairly simple analytic model of the aerial environment and drone dynamics and accounts for several limitations of real-life drones, e.g. acceleration limits, sensory noise, inaccurate rotor output, communication bandwidth, communication range, etc. The framework is optimized for the speedy execution of multi-agent simulations while it also provides a minimalist but handy GUI for the visual evaluation of the algorithms. The simulations were executed under Linux, either on a general desktop machine or - when thousands of parallel simulations were needed during evolutionary optimization - on the Atlasz supercomputer cluster of Eötvös University (\url{https://hpc.iig.elte.hu}). For details of the general features of the simulation framework see \parencite{Viragh2014} or check the actual open-source code at \url{https://github.com/csviragh/robotsim}. Supplementary Table S1 summarizes the used environmental parameters aimed at simulating realistic noises and dynamics.

Within this framework we defined a unique new traffic algorithm as described below, with model parameters summarized in Supplementary Table S2.

All the simulations presented in this paper were conducted through a very simple scheme of generating destinations for the agents. Every agent gets a random target point (independently) on the edge of a square-shaped arena of size $L$ x $L$. Once an agent reaches its destination it gets a new target, randomly, somewhere along another edge. New targets closer than $L/3$ to the current target are excluded to partially reduce increased density of agents at the sides. For further considerations regarding the inhomogeneous density distribution of agents, see Discussion.

When quantifying how good any traffic algorithm is, one needs to account for two aspects: $i$) how dangerous it is to navigate through the cloud of all the others $ii$) how quickly the agents reach their goal.

In our evaluation, we use collision risk as a quantitative measurement of danger. We consider two agents at under the direct risk of collision if they are closer than a given $r^\mathrm{coll}$ distance. The collision risk, $\psi^\mathrm{coll}$ is measured as the ratio of the number of collisions over the number of possible pairs of agents, averaged over simulation time length, $T$.

\begin{equation}
    \psi^\mathrm{coll} = \frac{1}{T} \int_0^T \frac{1}{N(N-1)} \sum_{i=1}^N\sum_{j \neq i} \theta \left(r^\mathrm{coll}-d_{ij}(t)\right),
    \label{eq_coll-risk}
\end{equation}

where $N$ is the number of agents, and $d_{ij}$ is the distance between agents $i$ and $j$.

To measure how quick agents can reach their targets, we first use the effective velocity.
We project the agent's current velocity to the line that would be its optimal linear route between its previous ($\mathbf{x}_i^{\mathrm{prev}}$) and next target ($\mathbf{x}_i^{\mathrm{next}}$):

\begin{equation}
    v^{\mathrm{eff}}_i(t)=
    \mathbf{v}_i(t)\cdot\mathcal{N}(\mathbf{x}_i^{\mathrm{next}}-\mathbf{x}_i^{\mathrm{prev}}),
    \label{eq_v-eff}
\end{equation}

where the operator $\mathcal{N}(.)$ normalizes its vector argument. The expression needs to be adjusted with a sign according to $\mathrm{sgn}\left((\mathbf{x}_i^{\mathrm{next}}-\mathbf{r}_i(t))(\mathbf{x}_i^{\mathrm{next}}-\mathbf{x}_i^{\mathrm{prev}})\right)$. Within the signum function is almost always positive except when an agent 'runs over' its current target (or starts backward from its previous one).

The effectiveness of the traffic flow, $v^\mathrm{eff}$, is given by averaging $v^\mathrm{eff}_i(t)$ over agents and time. Still, this is the average of a fundamentally local quantity, akin to particle flux density.

Thus, for a bigger picture, we also use a global quantity to measure the traffic flow. That is the throughput -- e.g. the number of agents arriving to their goals within a given unit of time. This can be derived as:

\begin{equation}
 \Phi^\mathrm{through} = \frac{v^\mathrm{eff}}{<l>} N,
 \label{eq_flux}
\end{equation}

where $<l>$ is the average length of direct paths between targets and $N$ is the number of agents. (An agent reaches a target in about every $<l>$/$v^\mathrm{eff}$ seconds, and there are $N$ of them.) The average direct path length must be measured separately in any scenario for assigning targets. For the scenario we use in every simulation (square-shaped $L$ x $L$ arena, with random targets on the edge, without allowing a path that is shorter than $L$/3), this gives $<l> = 0.947 L$. For a scenario where the new target point is randomly drawn from a circle of radius $R$, which scenario is used during the real drone experiments: $<l> = 4R/\pi$. For details on how to acquire these constants, refer to Supplementary Text S1. Supplementary Table S2 contains the used model parameters.

Our first result (Fig. \ref{fig_traffic-100A-flux}) shows that the new algorithm is able to govern the traffic of 100 autonomous drones with homogeneous desired speed in a reasonably wide density range. According to Fig. \ref{fig_traffic-100A-flux}, if 100 agents are using an arena of 275 m x 275 m, their traffic flow with maximum speed $v^{\mathrm{SPP}}$ = 8 m/s bears a tolerably low collision risk while the overall throughput is optimal. 

By "tolerably low risk" we mean that collisions are rare during a 10-minute experiment (for comparison, according to Eq. \ref{eq_coll-risk} collision risk of 1e-6 means 6 occurrences of agents below $R^\mathrm{danger}$ with 100 agents over 10 minutes). Note that reducing collision risk and increasing traffic flow are contradictory requirements thus satisfying both at the same time must be a tradeoff. In the presented results we have deliberately chosen a parameter setup where this tradeoff is visible and collisions are not "under-saturated". 
Secondly, since in real outdoor experiments (and in the simulations imitating those) stochastic noise is present, it is principally impossible to declare zero-risk guarantees. However, according to Fig. \ref{fig_traffic-100A-flux}, one can always choose the density of agents so that the collision risk is tolerable for the given application in terms of the price of a single drone or the possible damage caused by a collision. Furthermore, if safety is a concern in real-life applications, the way to further eliminate collision risk is through increasing the distance of avoidance and repulsion between agents. This of course comes at a price of reduced overall efficiency.

Looking at a null model where agents do not interact with each other (Fig. \ref{fig_traffic-100A-flux}a), we can say that the traffic algorithm presented here reduces collision risk by \emph{more than three orders of magnitude} compared to how many dangerous route crossings would occur by the sheer geometry of the traffic task. (At 27.5 m of mean free path the collision risk is reduced by a ratio of roughly 1:2500.)

An arena size of 275 m corresponds to 27.5 m of mean free path for 100 agents (considering a simplified connection: $MFP = L / \sqrt{N}$). To put this number in context one can compare it to the distance at which two frontally moving agents can stop with late reaction and limited deceleration. This stopping distance is only slightly smaller, 26.6 m for our realistic simulation case ($v^{\mathrm{SPP}}$ = 8 m/s, $t_{\mathrm{delay}}$ = 1 s, $a_{\mathrm{max}}$ = 6 m/s$^2$).

As Fig. \ref{fig_traffic-100A-flux} shows, providing more space for the agents results in ever-increasing effective velocity and practically zero risk of collisions, but flux (which is a group level measure) is decreasing since the speed of the agents cannot compensate for their sparseness.
The maximum flux of 1.961 s$^{-1}$ is over a double-fold gain compared to 0.96 s$^{-1}$ of our previous results \parencite{Balazs2018CoordinatedDrones} without any extra cost in terms of collision risk. 
Furthermore, the algorithm presented by this work is demonstrated to govern significantly more agents both in simulation (5000 vs 100) and in reality (100 vs 30). 

\subsection*{Scalability}

The algorithm is scalable by design. Each drone uses only local information (position, velocity and target of neighbours) to calculate its momentary preferred velocity vector, and this calculation is decentralized: performed on board by every drone. 
To examine whether scalability follows from these principles, we performed consecutive simulations with up to 5000 agents. For comparability, we kept the homogeneous desired speed and the mean free path (and thus density) constant. The mean free path was set to 27.5 m, which corresponds to the maximal flux for 100 agents.

\begin{figure}[!htb]
    \centering
    \includegraphics[width=0.75\columnwidth]{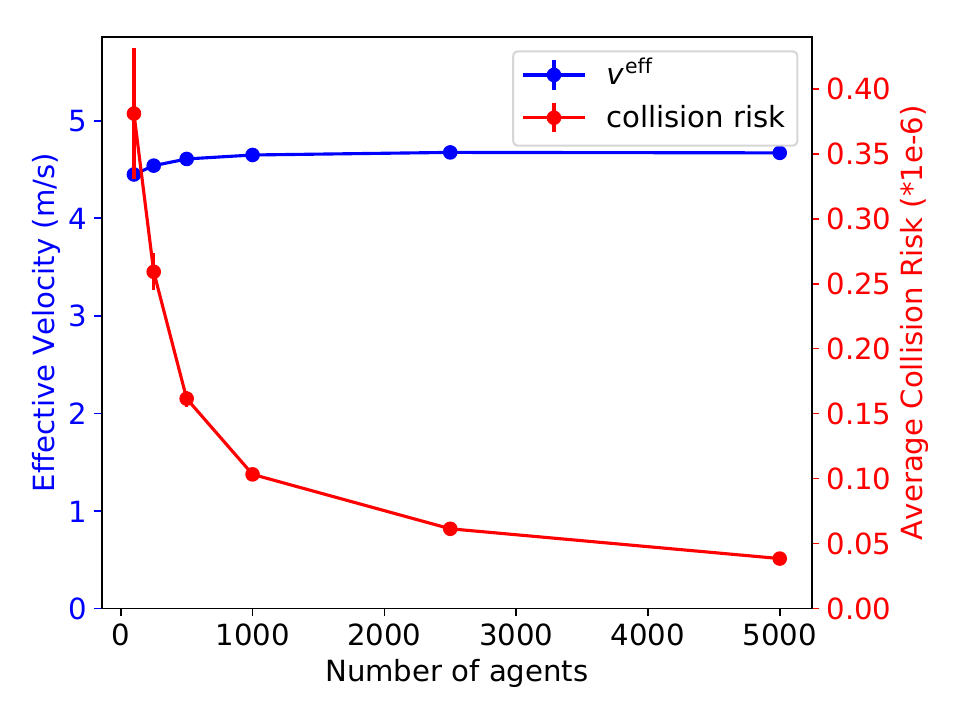}
    \caption{Scalability of traffic flow. Agents reach almost exactly the same effective velocity -- and the same throughput due to fixed density -- across orders of magnitude of change in their number. The collision risk, as the ratio of actual and all possible collisions is decreasing as more agents are involved, which corresponds to a somewhat increasing collision number compared to linear scaling with agent number. One hundred 10-minute simulations were averaged at each measurement point. Mean free path is 27.5 m in each setup (equals to $13.2*10^{-4}$ agents/m$^{2}$ density).
    Standard deviations are shown with error bars, but most often are too small to be visible.}
    \label{fig_traffic-flux-scale}
\end{figure}

As Fig. \ref{fig_traffic-flux-scale} shows, all the overall properties of the traffic flow are left practically unchanged even with the massively rising number of agents participating (same effective velocity and similar order in the number of collisions). Note that scaling the system up to even more agents has no extra local computational cost: the radius of interaction of the agents is limited locally, so involving more agents does not increase onboard computational load or the complexity of local decisions. One limiting factor could be the tolerable collision risk. Even though the collision risk is decreasing in larger systems, the number of collisions (based on the definition of collision risk) is increasing somewhat over linear scaling with agent number (around 270 times more collisions with 50 times more agents at similar density in the comparison between 100 and 5000 agents). If needed, of course this can be reduced to safer levels at the price of decreased throughput. 

Supplementary Movie S1 demonstrates the realistic 2D traffic with 5, 50, 500 and 5000 simulated agents.

Besides scalability in the number of agents, scalability in the velocity space is also possible, as interaction terms count with the finite acceleration of agents.
This means that the same algorithmic rules could apply for high-speed vehicles, and they will inherently start to avoid each other and reduce their velocity difference at a larger distance. Of course, other parameters of the algorithm, such as the range of repulsion or the range of avoidance can also be raised easily if higher speed ranges require additional safety measures. However, agents with higher speed \textit{must} be equipped with larger-distance communication modules to be able to detect others within their increased stopping distance.
For a figure very similar to Fig. \ref{fig_traffic-100A-flux}, hence demonstrating the scalability at a double-fold speed (16 m/s), refer to Supplementary Figure S1.

\subsection*{Heterogeneous traffic}

In practical applications it is very important to be able to treat heterogeneous agents in coordinated traffic. Our solution can handle both fundamental sources of heterogeneity: i) differences in dynamic properties (e.g., speed, acceleration); ii) differences in priority (i.e., hierarchical traffic).

\subsubsection*{Heterogeneity in agent dynamics}

Considering the previous result that the solution can be scaled up in the velocity space well due to the dynamic model of the algorithm, heterogeneity in the desired speed of agents becomes a triviality. In fact, it is equivalent of simply slowing a fraction of the agents down from the maximum speed (analogy: cars are not allowed to exceed the speed limit but can always go slower on the road). 

If physical or dynamic properties of agents (e.g., velocity, acceleration, size etc.) differ, the only important thing is to communicate these locally to each other to be able to take them into account when predicting the motion of others. Keeping ourselves to the road traffic analogy, this is similar to visually observing and understanding different dynamics of different vehicle types on the road (e.g. cars, buses, trucks, chariots, bicycles).

Supplementary Movie S2 shows the realistic simulation of a traffic scenario with 30 agents, where traffic conflicts are resolved smoothly when a different travel speed is assigned to each agent (linearly from the 2-32 m/s range). The typical effective velocities for the agents based on 100 simulations are shown in Supplementary Figure S2.

\subsubsection*{Hierarchical priorities}

In practical applications it is very important to be able to treat priorities of vehicles during traffic conflicts (consider e.g. the case of ambulance cars or fire trucks). We have achieved this formally by making the radius of avoidance of agents different.

Hierarchy is defined in a pairwise fashion, with time-independence and binary relations for the sake of simplicity. In each pair of agents we can prioritize one of them, or let the two be equal. These relations need to be prescribed previously and left fixed in this approach. 
(But as far as application on the drone fleet is concerned, it is easy to generalize to time-dependence if drones can agree on a method how to communicate and agree on new priorities).

Equality means that both agents apply the self-driving term to each other with the same $R^{\mathrm{avoid}}$ safety distance.
But if one of the agents is prioritized, it will not apply the prudent self-drive term with the normal $R^{\mathrm{avoid}}$ $\sim$ 12 m safety distance, but with a smaller $R^\mathrm{danger}$ $\sim$ 5 m danger distance, and it will also not apply friction interaction anyhow to its subdominant pair. Hierarchy leaves repulsion intact, because if the two really get closer than the repulsion distance, it means that due to some other limitations (inertia or lack of possibilities left open in the current traffic situation) the self-driving term could not deliver on its duties, so the primary interaction for safety needs to act.

Solidarity from the behalf of the boss in the given pair, however, is required to avoid catastrophes, after which both involved would feel indignant. 
The boss can only pay less attention (shrink the zone-to-avoid around the subordinate from $R^{\mathrm{avoid}}$ to $R^\mathrm{danger}$) if the collision course is not entirely his fault; in other words, the boss agent can only behave as such if the subordinate agent has a non-zero velocity component towards him. 
With this condition we can prevent situations when the subordinate and the boss are moving in the same direction, but the subordinate is slower, and the boss rushes into him from behind. 

Note that if everything goes as it should, the minimal distance of any two agents passing by each other will be the maximum of the distance they try to keep from each other. This is always $R^{\mathrm{avoid}}$! The fact that one (and never both) of them might use $R^\mathrm{danger}$ distance in the self-driving term merely encodes that the other, subordinate one will make much more of an effort to always keep the $R^{\mathrm{avoid}}$ distance.

To study the reaction of the system to pairwise priorities, we ran simulations of the strongest possible hierarchy compared to the egalitarian rules of avoidance. The strongest hierarchy, from a traffic point of view, is when agent 0 has priority over everyone else, agent 1 has priority over everyone except agent 0, agent 2 has priority over everyone expect agents 0 and 1, and so on. Equivalently: when the hierarchical graph is the complete directed acyclic graph, and the hierarchical matrix is an upper-triangle one.

\begin{figure}[!htb]
    \centering
    \includegraphics[width=0.75\columnwidth]{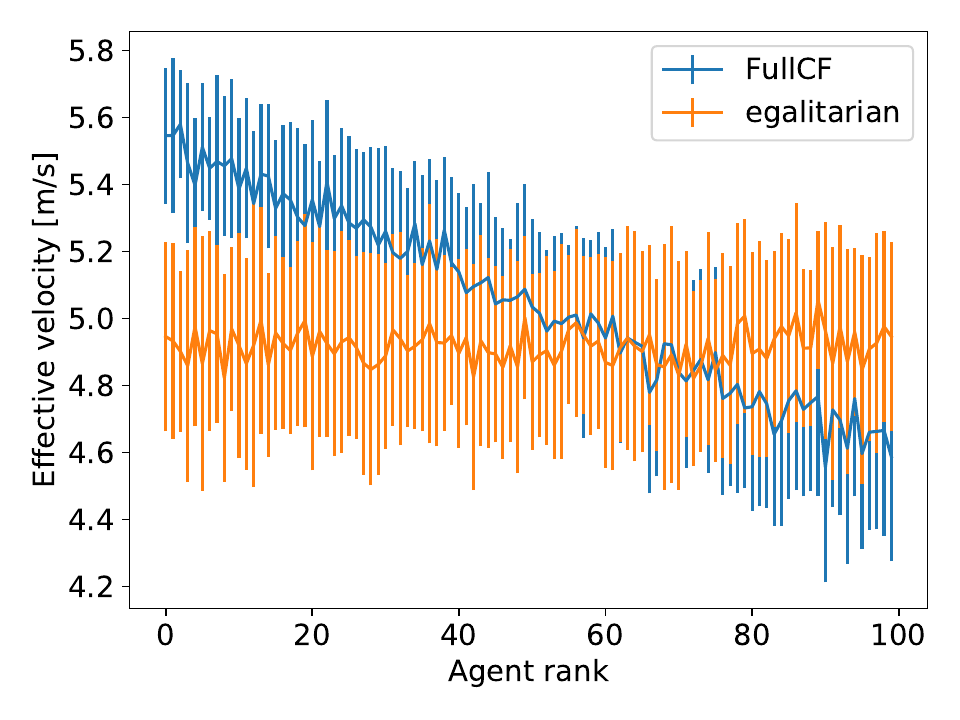}
    \caption{Effective velocity as a function of agent rank in the strongest hierarchy (full cycle-free, FullCF) and when no hierarchy is introduced (egalitarian). Agents with lower label-number are bossing a lot of their peers (e.g. rank 0 corresponds to the highest boss), which means they make avoiding maneuvers less frequently, which enhances their effectivity. Agents with a lot of bosses (higher ranks) are compensating this, hence are losing effectivity. Both curves are averaged from 100 simulations each, at desired speed 8 m/s.}
    \label{fig_traffic-hierarchy}
\end{figure}

Fig. \ref{fig_traffic-hierarchy} shows the expected effect of hierarchy when the desired travel speed is kept equal for all agents. Bosses are moving more directly towards their targets, while subordinate agents are doing more evasive maneuvers to keep the flow collision-less, hence are losing efficiency. 
Note that Fig. \ref{fig_traffic-hierarchy} shows the two extremes of the possible priority distributions: egalitarian and completely hierarchical cases. In practice any other required dominance scenario can be mixed as priority is a pairwise attribute.

Finally, note that in practical situations the heterogeneity of desired travel speed \textit{and} priority can both enhance the final asymmetry of agent effectiveness, so the effects of Fig. \ref{fig_traffic-hierarchy} and Supplementary Figure S2 can support each other.

\subsection*{Layered traffic in 3D}

Three dimensional space is very sparse. To challenge our traffic algorithm, we examined purely two dimensional (planar) cases first, as with a given number of agents it is easier to increase the density and thus the number of conflicts to resolve in 2D. However, when the final goal is to maximize flux, one can extend the two dimensional solution to the third dimension as well. For this, we introduced separate planar layers of traffic upon each other with a small but safe vertical distance between them and forced agents to choose from the layers based on their desired ideal direction of motion between the last and next target. 

The cost of having more layers is that it takes extra time and energy to switch layers vertically before heading to destination and at arrival. 

The benefit first of all is that the average density over a horizontal area in one layer will be reduced as more vertical space can be utilized above. Besides, with more layers the desired direction of the flow in each one will be more aligned and thus the possibility of frontal or lateral traffic conflicts will decrease.

In simulation we compared the case of 1, 3 and 5 layers, with a base layer and possible other layers above and below symmetrically. Since target points remained in the base layer, the trajectories were broken down into three stages: 1) reaching the layer of motion vertically; 2) moving towards the target horizontally; 3) going back to the base layer vertically. Note that $2*N$ layers have similar costs (time spent with elevation and descent) but less benefit (decrease of density) than $2*N+1$, thus even numbers of layers got excluded from the quantitative investigation.

When more than one layer is used, we still insist the target points be reached on only one of them: on the base layer. 
An agent having just reached its previous target at the base layer gets a new target randomly. Based on the direction of the straight route relative to the North and the number of layers, the agent knows which layer it needs to use to get to the next point. In the case of 3 layers for example, 0\degree-120\degree: layer 0, 120\degree-240\degree: layer 1, 240\degree-360\degree: layer -1.
Then it starts its elevation (or descent) as a separate phase from trafficking. Until the vertical position of the desired level is reached, $\mathbf{v}_i^\mathrm{self-drive}$ of Eq. \ref{eq_desired-velocity} is solely vertical. Repulsion and friction interactions are still working, being applied to any two agents that have a vertical distance less than $h^{\mathrm{layer}}\alpha^{\mathrm{layer-overlap}}$, where $h^{\mathrm{layer}}$ is the layer height difference (in our case: 10 metres) and $\alpha^{\mathrm{layer-overlap}}$ is the extent of overlapping of the layers (must be between 0 and 1, in our case: 0.5). This way, even if the entry point of an agent is somewhat crowded at the new layer, it is getting cleared up by the repulsion interaction. 

When the new layer is reached, the agent starts its 2D traffic movement, only taking into account agents that are closer than $h^{\mathrm{layer}}\alpha^{\mathrm{layer-overlap}}$ vertically. Once the target position is reached, agents ascend or descend to the base layer (in the same way as described above) to finalize their mission and get a new target. If the agent gets a bit pushed away from the target during vertical motion, it needs to re-reach the target on the base layer before getting a new one.

\begin{figure}[!htb]
    \centering
    \includegraphics[width=0.9\columnwidth]{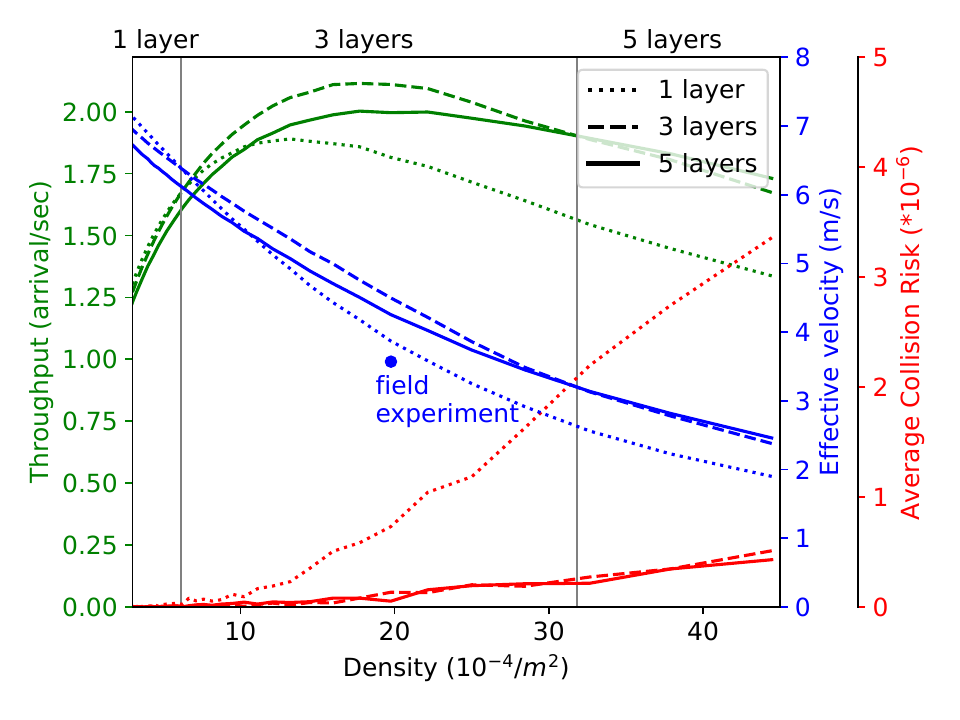}
    \caption{Fundamental traffic diagram comparing simulations with one (dotted lines), three (dashed lines) and five (solid lines) layers. Agents select different layers depending on their direction of velocity. Figure shows the average effective velocity (blue), the average throughput (green) and the average collision risk (red) of 100 simulations at each density value. The optimal number of layers (i.e. the one with highest throughput at the given density section) is marked at the top of the figure. If density increases, more layers provide higher throughput. The throughput has maximum at middle density. The effective velocity value of the field experiment with 100 drones is also added to the plot for reality gap comparison. Simulation parameters: 100 drones, 8 m/s travel speed, square-shaped arena with side length between 150-600 m. Experiment parameters: 97 drones, 6 m/s travel speed, two layers, circular arena with 125 m radius. Note that the difference between the experimental and simulated data disappears once the exact same parameters are set: two layers, 97 agents, 6 m/s travel speed simulations give 3.52$\pm$0.02 m/s effective velocity as compared to 3.57 m/s in real flight. Different parameters for this figure are plotted because odd numbers of layers are shown to be more effective and 8 m/s is still quite a safe speed -- even though none of these two are exploited in our proof-of-concept field experiment.}
    \label{fig_traffic-flux-layers}
\end{figure}

Fig. \ref{fig_traffic-flux-layers} summarizes the simulation results for one, three and five layers, while Supplementary Movie S3 shows the actual simulated traffic for 1-4 layers.

\begin{figure}[!htb]
    \centering
    \includegraphics[width=0.9\textwidth]{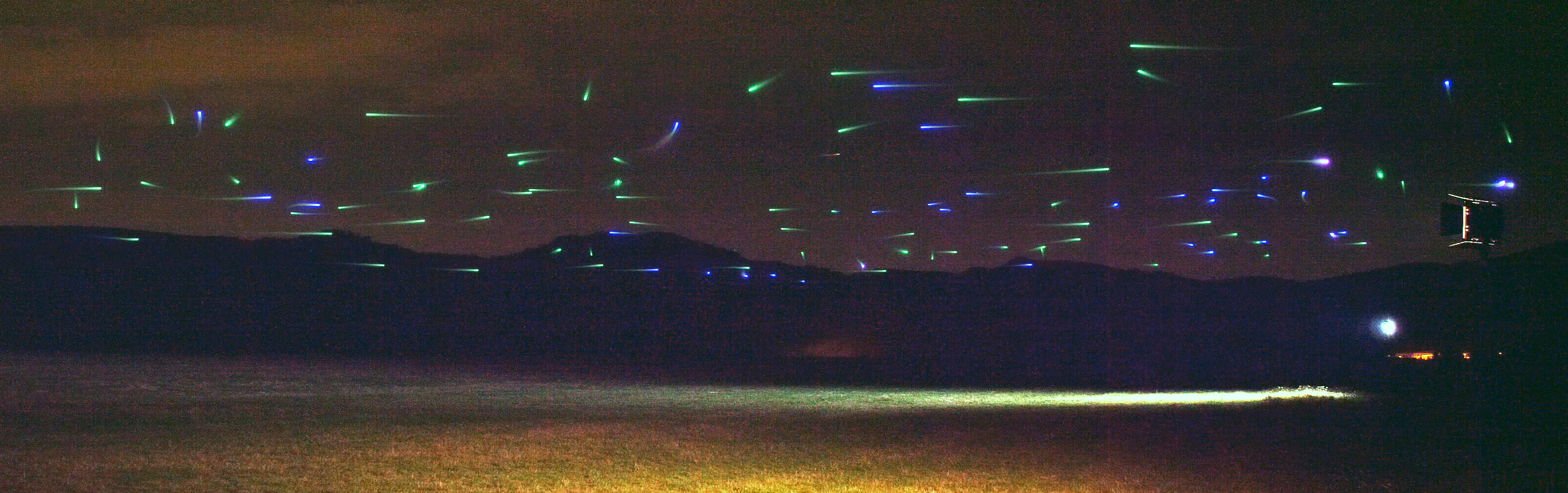}
    \caption{Post-processed image of the raw video footage of the field experiment with 100 drones performing autonomous decentralized traffic in two layers. The image was created iteratively, by adaptively averaging consecutive video frames into a background image, and then getting the pixel-level lighter color values from the current background and the new frame. The light indicator attached to each drone shows the status of the algorithm: green means free motion, blue means the drone is avoiding others, purple means the drone is queueing, white flash indicates that the target has been reached.}
    \label{fig_traffic-field-experiment}
\end{figure}

It is clear from the simulation results that raising the number of layers from 1 to 3 is beneficial when a single layer gets too crowded. Collision risk also decreases with the new layers introduced. However, it is also clear that raising the number of layers further adds less benefit than before. The main reason for this is that new layers do not raise the extent of the takeoff/landing areas, which still remain a bottleneck. For more thoughts about this geometrical limitation see Discussion.

\section*{Proof of concept with a drone swarm}

A field demonstration of the decentralized traffic algorithm has been conducted with a fleet of 100 drones. 

We used the drone fleet and the expertise in large-scale drone-swarm deployment of CollMot Robotics Ltd. These quadcopters use a PixHawk1 type autopilot with slightly customized ArduCopter code (available at \url{https://github.com/skybrush-io/ardupilot}), a U-blox M8P RTK-compatible GNSS receiver, an IBSS-compatible 2.4 GHz Odroid WiFi module 0 and a 433 MHz ISM-band SIK radio. The traffic algorithm was running on an on-board Odroid C1+ companion computer, which sent velocity control commands to the autopilot at 20 Hz.

RTK corrections were sent through both the WiFi and ISM bands from a ground station running Skybrush (\url{https://skybrush.io}). RTK corrections have been actively used throughout the experiment to increase positional accuracy, however, from the algorithm's structural point of view corrections are not required, they simply reduce the allowable minimum distance between drones during flight. Without RTK corrections one needs larger repulsion radius and stopping distance between the drones and thus in general a larger available flight area.

Drones communicated with each other through an ad-hoc (IBSS) WiFi network with non-directional (isotropic) antennas. In the network UDP status messages were broadcast at 10 Hz, containing a timestamp, one's ID, current position, velocity, target position and general health and administrative status. The low output power of the WiFi modules and the broadcast-type UDP protocol allows for good scaling of the communication network, as packets are typically received from only closeby neighbors (reception typically starts to degrade above ~80m, see a comprehensive analysis about the measured communication network quality in Fig8 of \parencite{Vasarhelyi2018}), hence the network never really gets flooded. Anyhow, the theoretical maximum reception bandwidth required for 100 drones, <100 byte status packets at 10 Hz is ~100 kByte/s, which can be handled easily by any state-of-the-art WiFi modules. 

During the experiments we required stable GPS reception and in general conditions for normal flight operation. The drones had a complex onboard fail-safe mechanism which -- depending on the severity of any detected error -- forced the drones to return to home at a safe altitude individually or land immediately, without endangering the safe operation of others. 

The experiment used two layers for the traffic spaced 14 m above each other. Travel speed was set to 6 m/s horizontally and 1.5 m/s vertically, therefore, the selected 10 Hz status update resulted in maximum 0.6 m positional ambiguity from neighbors, while timestamp comparisons were also used for a linear extrapolation of position since last received packets. This update frequency proved to be high enough in general for all our autonomous drone experiments in the past years as well, not just this one. Traffic targets were selected randomly from the edge of a circle with 125 m radius. Note that using a circle instead of a rectangle (such as in the presented simulations) had no specific reason; it was an instantaneous choice for the sake of diversity as according to (not presented) simulations it has no major effect on the outcome. Other parameters of the algorithm are fairly similar to the optimized simulations (see Supplementary Table S2 for the detailed parameter setup).

Fig. \ref{fig_traffic-field-experiment} shows a section of the raw video footage of the field experiment blended into a single image.

Supplementary Movie S4 is our video abstract that also contains video footage about the fleet performing decentralized self-organizing traffic. Supplementary Movie S5 is a three-minute sample from the experimental tape. 

An interactive visualization of the recorded flight logs (97 out of 100 could be retrieved) is also available at \url{https://share.skybrush.io/s/traffic-layers/}.

\begin{figure}[!htb]
    \centering
    \includegraphics[width=0.85\columnwidth]{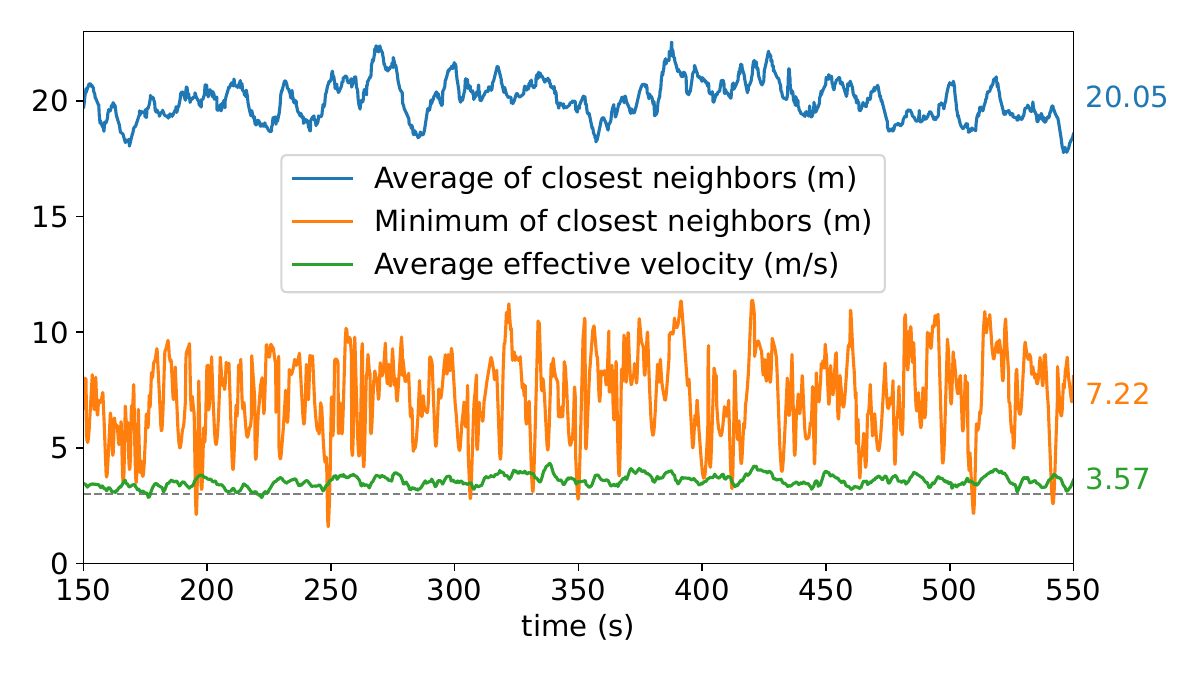}
    \caption{Statistical output based on a 400 s stationary section of the flight logs of 97 drones performing dense self-organized traffic. The y axis of the figure is common for all quantities numerically, while the unit of each one is shown in the figure legend. The absolute average of all quantities is marked at the right side of the figure with the same color as the corresponding timeline. The 3 m collision risk distance used in simulations is also plotted with a gray dashed line for reference. Even though the 3 m safety distance is violated a few times, drones never got closer to each other than 1.5 m, and the results show an overall stable and efficient trafficking motion.}
    \label{fig_real-flight-timeline}
\end{figure}

Fig. \ref{fig_real-flight-timeline} shows statistical output from the available real flight logs. The results show stable and efficient traffic behaviour, and we confirm that the whole flight was performed without collisions. Nevertheless, if we or the readers ever need to repeat such a complex experiment, it is advised to take into account more reality gap and raise the radius of avoidance to completely eliminate the occasional dangerous spikes in the minimum distance between drones.

\section*{Discussion}

We have presented a decentralized control strategy for many cooperating autonomous agents with independent tasks that require coordination and continuous conflict resolution while moving in a common airspace.

Our system has been designed with all the conceptual guiding rules and benefits of self-organization in mind, as described in \parencite{Gershenson2020GuidingSystems}: i) it is inherently \textit{adaptive} and \textit{robust} due to its distributed nature; ii) it shows signs of \textit{antifragility} in frontal pairwise conflicts where positional noise facilitates decisions about which direction to avoid each other; iii) our altruistic traffic rules act as efficient \textit{mediators} to minimize pairwise conflicts in the traffic flow; iv) polite queueing behaviour builds on the \textit{slower is faster} effect; and finally, v) \textit{heterogeneity} is treated thoroughly to provide a solid base for future emergent complexity of autonomous traffic.

The communication demand of our traffic algorithm is relatively low: agents require only local information to be received from interacting neighbors, such as their position, velocity, and most importantly, their \textit{destination}. This last piece of active information as part of the situational awareness of agents is crucial for achieving a certain efficiency; without this agents would not be capable of predicting each others motion well enough and their reaction to the motion of others would be slower and less accurate. This message is perfectly in line with the results of our latest article about flocking \parencite{Balazs2020AdaptiveFlocking}, where actively communicating one's own \textit{will} turned out to be the key for sidestepping the strict tradeoff between persistance/stability and responsivity/reactivity.

The primary advantage of our decentralized approach is that the overall self-organized traffic flow will not only be efficient and stable, but also completely \textit{scalable} to masses of vehicles. This comes in a package with easy implementation: our algorithm is also very easy to program to real drones and requires only local low power communication infrastructure. These two facts together make our approach a good candidate to aid autonomous flights, or to become even part of a global standard in decentralized Unmanned Aircraft System Traffic Management.

Regarding real-life applications, our solution also easily extends to vehicles with diverse dynamic properties, such as their acceleration or speed range, while it can also handle any possible (static or dynamic) priority distribution or hierarchy within the coordinated airspace. This latter property is actually equivalent to handling size heterogeneity with simply changing the radius of avoidance at the individual's level. All in all, handling heterogeneity and priorities of agents is very helpful for the interoperatibility between different UAV systems and for future UTM + ATM integration purposes.

Our solution for multi-UAV coordination is first presented for the more dense case of strictly two-dimensional traffic. It is also extended to a layered three-dimensional structure, which consumes very small vertical space (suitable for the typically low-altitude UAV flights), enhances the overall efficiency, but interestingly, not as much as we predicted beforehand. The reason for this can be revealed by investigating the nature of bottlenecks when large number of drones commute between pre-assigned positions. The decrease of efficiency here is rooted more in the congestion at the departure/arrival zone than out in open-air space -- the destination points are along a lower dimensional part of space (in our case line segments) than the quasi-two-dimensional area in which the drones move away from the terminals. In fact, this evidence is also observable in i) one-dimensional road traffic, where the usual constraint is arrival to central points with zero dimension (e.g. entry points of a city); ii) two-dimensional human traffic in conference halls while queueing for food along one-dimensional outlines of tables; iii) current air traffic, where the throughput of airports is a stronger bottleneck than airspace availability above. This observation draws the attention to the need of careful planning of future smart cities with dense unmanned aerial traffic requirements, where traffic throughput will only be able to expand above a certain limit if unmanned "airports" will use the third dimension also (e.g., having equivalent, well-separated landing spots on all floors of high buildings). 

In the present article, we focused on the challenges arising in open-air traffic, where drones avoid each other based on absolute position data actively transmitted from other agents. Several practical applications -- such as smart cities -- shall demand general passive obstacle avoidance capabilities as well for resolving traffic in confined spaces. Even though this direction lays outside the scope of the present paper, once drones get equipped with proper sensors to detect static objects around them \parencite{Soria2021PredictiveEnvironments}, the methods presented here shall provide a solid base for handling general object avoidance as well as a special case (static objects are virtually agents with zero speed and in the simplest case different radius of avoidance). This could very well be the focus of future research.

There are also several options for extending our algorithm to non-layered 3D motion. A quick extension is to keep all interaction terms in 2D and force interactions for all agents that are close either horizontally or vertically (this is what actually happens already at targets, while changing layers). Another direction could be to generalize all interaction terms for a full 3D case. This is straightforward for repulsion and friction but not that trivial for the self-drive term, and also brings in questionable requirements for the dynamics from a practical point of view (raising altitude consumes more energy in the air due to gravity and the gained potential energy cannot be recovered during descent).

Last but not least, our decentralized traffic algorithm is currently optimized for multi-rotor type vehicles, which is demonstrated on the field with a swarm of 100 autonomous quadcopters. If needed, the philosophy of the approach and a substantial portion of the algorithm can be used for fixed wing aircraft as well. However, one has to be very cautious about how to keep the fixed-wing's speed about its stall speed. This surely requires modifications in the queueing behaviour (e.g., circular loitering instead of hovering) and adds extra restrictions to the instantaneous selection of the self-drive term.

The range of applicability of our results is wide. Aerial missions with more than a single drone involved require coordination between the cooperating agents, which can be solved now with our algorithm. As mentioned above, an extended version of our collision avoidance strategy can also be part of a global UTM solution if autonomous agents get equipped with the proper communication devices and onboard DAA control units running the algorithm. But due to its low computational cost, the decentralized traffic model can also be used in a centralized way in air traffic management as an automated planner or recommendation system for human air traffic controllers.

\section*{Declarations}

{\bf Acknowledgments: }The authors are grateful for all helpers at drone experiments, namely to Evelin Berekméri, Gáspár Hegyi, Göksel Keskin, Pedro Correa Pereira Vasco de Lacerda, Máté Nagy, András Tekus, Gergely Vadász and his family, Csilla Vitályos and András Zabó. 
We also thank for the flawless grant administration for Anikó Farkas and Mária Kolozsvári.
Drone fleet and Skybrush fleet control software was provided by CollMot Robotics LLC.

{\bf Author contributions:} The project was concieved by GV. Authors BB, GV and TV were developing the concept and the algorithms, BB and GV were developing the simulation, TN and GV were developing the software for the drones, GS, GV and TN were developing the drone hardware, BB, TN, GS and GV were performing the drone experiments. The paper was mostly written by BB, GV and TV, while it was thoroughly reviewed by TN and GS.

{\bf Competing interests:} All authors declare that they have no competing interests.

{\bf Availability of data and material:} Source code of our generic robotic simulation framework is available at \url{https://github.com/csviragh/robotsim}.

{\bf Funding: }This work was partly supported by the following grants:
MTA-ELTE Statistical and Biological Physics Research Group,
K\_16 Research Grant of the Hungarian National Research, Development and Innovation Office, Grant No: K 119467,
US AF Office of Scientific Research, Grant No: FA9550-17-1-0037, SNN\_21 Research Grant of the Hungarian Ministry of Innovation and Technology, Grant No: 139598.
National Research, Development and Innovation Office – NKFIH: OTKA SNN 139598

{\bf Supplementary Material:}
\begin{itemize}
  \item Movie S1. Realistic simulation of 2D decentralized drone traffic with 5, 50, 500 and 5000 agents.
  \item Movie S2. Realistic simulation of 2D decentralized drone traffic with heterogeneous travel speed (2-32 m/s).
  \item Movie S3. Realistic simulation of 3D decentralized drone traffic with 500 drones in 1, 2, 3 and 4 layers.
  \item Movie S4. Summarizing documentary (video abstract) with simulation, flight log visualization, and footage on real flights.
  \item Table S1. Environmental parameters used in the realistic simulations.
  \item Table S2. Model parameter setup of the simulations and the drone field experiment.
  \item Figure S1. Traffic flow statistics of 100 agents at 16 m/s desired travel speed, as a function of density.
  \item Figure S2. Traffic flow statistics of 30 agents with heterogeneous desired speed between 2-32 m/s.
  \item Text S1. Geometrical constants for acquiring throughput.
  \item Flight log visualizations are available at \url{https://share.skybrush.io/s/traffic-layers/}
  \item Source code of our generic robotic simulation framework is available at \url{https://github.com/csviragh/robotsim}.
\end{itemize}

{\bf Ethics approval and consent to participate:} Not applicable.
{\bf Consent for publication:} Not applicable.


\printbibliography

\end{document}